\documentclass{article}

\usepackage{arxiv}

\usepackage[utf8]{inputenc} 
\usepackage[T1]{fontenc}    
\usepackage{hyperref}       
\usepackage{url}            
\usepackage{booktabs}       
\usepackage{amsfonts}       
\usepackage{nicefrac}       
\usepackage{microtype}      
\usepackage{lipsum}		
\usepackage{graphicx}
\usepackage{doi}

\usepackage{framed,multirow}
\usepackage{amsmath}
\usepackage{color,soul}
\usepackage[numbers]{natbib}

\title{Face sketch to photo translation using generative adversarial networks}

\date{} 					

\author{ Nastaran Moradzadeh Farid \\
Department of Computer Engineering\\
	Amirkabir University of Technology\\
	Tehran, Iran \\
\And
Maryam Saeedi Fard\\
Department of Computer Engineering\\
	Amirkabir University of Technology\\
	Tehran, Iran \\
\And
Ahmad Nickabadi\\
Department of Computer Engineering\\
	Amirkabir University of Technology\\
	Tehran, Iran \\
	\texttt{nickabadi@aut.ac.ir}
}

\renewcommand{\shorttitle}{\textit{arXiv} Template}

\hypersetup{
pdftitle={A template for the arxiv style},
pdfsubject={q-bio.NC, q-bio.QM},
pdfauthor={David S.~Hippocampus, Elias D.~Striatum},
pdfkeywords={First keyword, Second keyword, More},
}

\begin{document}
\maketitle

\begin{abstract}
	Translating face sketches to photo-realistic faces is an interesting and essential task in many applications like law enforcement and the digital entertainment industry. One of the most important challenges of this task is the inherent differences between the sketch and the real image such as the lack of color and details of the skin tissue in the sketch. With the advent of adversarial generative models, an increasing number of methods have been proposed for sketch-to-image synthesis. However, these models still suffer from limitations such as the large number of paired data required for training, the low resolution of the produced images, or the unrealistic appearance of the generated images. In this paper, we propose a method for converting an input facial sketch to a colorful photo without the need for any paired dataset. To do so, we use a pre-trained face photo generating model to synthesize high-quality natural face photos and employ an optimization procedure to keep high-fidelity to the input sketch. We train a network to map the facial features extracted from the input sketch to a vector in the latent space of the face generating model. Also, we study different optimization criteria and compare the results of the proposed model with those of the state-of-the-art models quantitatively and qualitatively. The proposed model achieved 0.655 in the SSIM index and 97.59\% rank-1 face recognition rate with higher quality of the produced images.
\end{abstract}


\section{Introduction}
\label{sec:Introduction}

Converting black and white facial sketches into realistic colored photos is one of the most interesting and practical issues in the field of image processing and machine vision that have been offered in various models in recent years. The sketch is referred to as the first drawing or art design that has the basic features of a face. In sketches, the details of the face, including the face's outline and color are rendered inaccurate and gray-scale.

Facial sketches can be classified into three types \cite{barbadekar2016survey}: 1) viewed sketches, 2) forensic sketches, and 3) composite sketches. Viewed sketches are drawn by the human hand while viewing the photograph. Forensic sketches are facial sketches drawn by artists based on the description provided by a third-party person. Composite sketches are facial sketches created using software kits that allow an operator to select different facial components.

Forensic and composite sketches are usually used for face recognition.
Face recognition from sketches has been a challenging task for a while and different methods have been developed for it \cite{klum2013sketch,wang2008face,tang2004face,tang2003face}.
There are lots of difficulties in sketch-based face recognition since photographs and sketches belong to two different modalities \cite{barbadekar2016survey} with completely different structural and appearance features. Moreover, the patches drawn by a pencil on paper have different textures from the human skin captured in a photo. So, it will not be easy to match sketches with realistic photos. One solution to these difficulties is to convert the sketch to photograph or vice versa and then compare the faces in the same space.

During the past decades, several methods have been proposed for creating colored realistic images from sketches. Some of these models just add color to the input image while the others follow a two-step model in which the input sketch is first translated into a descriptive feature vector and then the final image is synthesized using this vector.

Recently, GAN models have shown great performance in creating real face photos \cite{karras2019style,karras2017progressive}. In this paper, we propose a general framework for generating high resolution and high-quality colorful images from the face sketches. In this model, a face generating GAN is used for converting an intermediate latent vector to the final photo. The overall task of the other components of the proposed model is to find the most appropriate latent vector. To do so, the high-level characteristics of the input face sketch are first extracted in the form of a feature vector which is then mapped into the latent space of the face generating GAN. After this initialization step, an optimization step is adopted to increase the similarity of the synthesized photo to the sketch. This step iteratively adjusts the GAN’s latent vector, based on several defined optimization criteria to preserve the quality and fidelity of the produced face photo. It should be noted that the proposed model does not need to be trained on paired sketch-photo data and the input sketch does not need to belong to the training dataset of the model. The experimental results show the superiority of the proposed model both in qualitative and quantitative measures.

In Section \ref{sec:related_works}, we mention some related works. After that, in Section \ref{sec:Proposed_Framework}, we explain the proposed method to convert a face sketch to a real photo. The experimental results are reported in Section \ref{sec:Experiments}. Finally, Section \ref{sec:conclusion} concludes this paper.

\section{Related Works}
\label{sec:related_works}

The proposed models for sketch to photo conversion can be divided into two main groups: before-GANs (or \emph{shallow learning}-based) methods and GANs-based (or \emph{deep learning}-based) methods \cite{zhang2019cross}, the latter utilizes deep neural networks for image synthesis while the former employs other machine learning techniques. In the following subsections, the trends and prominent works of each category are reviewed.

\subsection{Before-GANs methods}
\label{subsec:before_GANs}

The sketch to photo translation methods before GANs can be categorized based on their model construction techniques to: 1) subspace learning-based approaches, 2) sparse representation-based approaches, and 3) Bayesian inference-based approaches \cite{wang2014comprehensive}.

Subspace learning refers to the technique of finding a subspace $R^m$ embedded in a high dimensional space $R^n$ ($n > m$). As examples of this approach, \cite{zhang2011face} propose a model based on support vector regression which handles the difficulty of losing vital details and \cite{wang2013transductive} present a method which incorporates the given test samples into the learning process and optimizes the performance on these test samples.

Sparse representation is a decomposition technique that represents a signal $y_{sig} \in R^n$ into a linear combination of basic signals $D_i \in R^n (i = 1, ... , k)$, weighted by nonzero coefficients. As a sparse representation-based method, \cite{wang2011face} propose a multi-dictionary sparse representation based model for high-quality sketch-photo synthesis that uses locally linear embedding to estimate an initial sketch or photo. \cite{gao2012face} propose another automatic sketch synthesis algorithm based on sparse representation. Their method works at patch level and is made out of selecting sparse neighbors for an underlying estimation of pseudo-images and sparse-representation-based improvement for additional improving the image quality.

In Bayesian inference-based approaches the output is obtained from Bayes’ theorem and the maximum a posteriori (MAP) decision rule. As a Bayesian inference-based method, \cite{peng2015multiple} propose a multiple representation-based face sketch-photo-synthesis method that adaptively combines multiple representations to represent an image patch. Then, Markov networks are utilized to take advantage of the relationships between adjacent patches.

\subsection{GANs-based methods}

\label{subsec:GAN_based}

With the invention of GANs \cite{goodfellow2014generative}, generative models have been successfully employed in many image generation and image to image translation tasks. Recently, several GANs-based methods have been proposed for sketch to photo translation which can be categorized as 1) image to image translation methods, 2) face photo-sketch synthesis methods, and 3) methods using pre-trained GANs.

\textbf{Image to image translation methods:} The image to image translation methods are general models and encompass techniques for any cross-domain image translation including face photo-sketch conversion. Two main branches of this category are the supervised and unsupervised methods \cite{yongxinsurvey}. The goal of the first branch is to learn the mapping between an input image and an output image using a training set of aligned image pairs \cite{yongxinsurvey}. \cite{isola2017image} proposed the supervised Pix2Pix model for several tasks such as labels to street scenes and image colorization.

In unsupervised image to image translation, source and target image sets are completely independent with no paired examples between the two domains. Because acquiring paired training data is expensive or sometimes impossible \cite{yongxinsurvey}, \cite{zhu2017unpaired} proposed CycleGAN for the unsupervised image to image translation tasks which have been shown to be successful in many applications where paired data is not available. However, failures occur due to the distribution imposed by the training dataset. Similar to CycleGAN, \cite{yi2017dualgan} proposed unsupervised DualGAN method which can generate results comparable to or even better than the results of supervised Pix2Pix \cite{isola2017image} model. In other work, \cite{wang2018high} proposed a method based on the CycleGAN idea and employed multi adversarial networks for facial photo-sketch synthesis. In another study, \cite{chao2019high} presented a high fidelity face sketch-photo synthesis using a deep neural U-net as generator and Patch-GAN with residual blocks as discriminator.

DRIT++ \cite{lee2020drit++} is another image-to-image translation model that generates diverse outputs without paired training images. As stated in this work, generating multiple possible outputs from a single input image is one of the main challenges of the image-to-image translation task. So, this model disentangles the latent space into a content space and a domain-specific attribute space. The former encodes common information between domains, and the latter can model the distinct varieties given the same content. Also, a content discriminator is applied to ease the representation disentanglement. 
Moreover, Council-GAN \cite{nizan2020breaking} as an image-to-image translation framework, instead of using a cycle, is based on collegiality between GANs. Precisely, rather than utilizing a pair of generator/discriminator architecture, it uses a group of triplets (one generator and two discriminators for each member), i.e. the council, and improves the variety of the generators' results. This idea makes domain transfer more stable and diverse.

\textbf{Face photo-sketch synthesis methods:} In these methods, the facial sketch is directly translated into the corresponding real face photo or vice versa.  The methods of this category concentrate only on the facial sketch and photo modalities. \cite{osahor2020quality} propose a GAN which synthesizes multiple outputs from the input sketch with different attributes like hair color. They use multiple loss functions to preserve identity and different attributes. Similarly, \cite{lin2020identity} propose an adversarial model for sketch-photo synthesis, which contains one generator and two discriminators for ensuring image quality and identity preservation. \cite{zheng2019feature} propose a model for both image to sketch and sketch to image conversion that uses Cycle-GAN as its baseline. They use a feature auto-encoder to refine the synthesis results. \cite{zhu2019deep} propose a framework for bidirectional photo-sketch mapping. They use a middle latent domain $Z$ between photo and sketch domains, and multiple loss functions to learn mappings.

Moreover, recently, \cite{yu2020toward} proposed Composition-Aided Generative Adversarial Network (CA-GAN). In this model, the paired inputs consisting of a face photo/sketch and the corresponding pixel-wise face labeling mask are utilized to generate the portrait. Furthermore, an improved reconstruction loss and a perceptual loss based on a pre-trained face recognition network are used. Finally, the developed form of CA-GAN named stacked CA-GANs (SCA-GAN) is presented for refinement. As this framework has used both the image appearance space and structural composition space, the generated face photos and sketches are natural. Also, \cite{fang2020identity} proposed an Identity-Aware Cycle Generative Adversarial Network (IACycleGAN) which concentrated on both face photo-sketch synthesis and recognition. So, to this purpose recognition networks are utilized in this model and are first fine-tuned by the generated images from CycleGAN, and also a perceptual loss is used to regularize the performance of the IACycleGAN. 

Also, Chen et al. in \cite{chen2020deepfacedrawing} proposed the DeepFaceDrawing model which can translate hand-drawn face sketches to real photos. They learned feature embedding of key face components and used a network to map the embedded features to the real photo. Furthermore, the DeepFacePencil \cite{li2020deepfacepencil} is a sketch-based face image synthesis framework based on hand-drawn sketches. The dual generator training strategy and spatial attention pooling (SAP) module are used in this model. The SAP module adjusts the spatially varying balance among the image naturalness and the conformance within the sketch and the synthesized image.

\textbf{Methods using pre-trained GANs:} In these methods, an existing model previously trained on another dataset with the ability of generating high-quality face images is used to translate a sketch to the corresponding photo. In this approach, first, the input sketch is mapped to the network’s latent space and then changes are applied to make the generated image closer to the input sketch. The main advantages of this approach are that creating the sketch to photo model does not require heavy training on large datasets and the model can be easily adapted to changes in the objective (loss) function. This approach has been used in face editing \cite{collins2020editing,shen2020interpreting} and text-to-image translation tasks \cite{wang2020faces} .

As an example, Pixel2style2pixel \cite{richardson2021encoding} is introduced based on the pretrained StyleGAN generator. It includes an encoder architecture and its main goal is to encode an arbitrary image directly into the intermediate latent space of StyleGAN. Subsequently, latent space manipulation is possible to edit the real image. The Pixel2style2pixel model is used in the sketch to photo conversion with the difference that the utilized sketches are incomplete freehand. 
       
Similar to Pixel2style2pixel, lots of works have focused on StyleGAN and intend to use the power of this model for better image synthesis and manipulation. For this purpose, various methods have been proposed to invert an image into StyleGAN's latent space. For example, \cite{abdal2019image2stylegan,karras2020analyzing} proposed methods to encode an input image into the latent space of StyleGAN. \cite{abdal2020image2stylegan++} developed previous methods and used them for image editing. Also, \cite{tewari2020pie} used a hierarchical optimization to embed an image-in-the-wild to the styleGAN's latent space and then performed editing via resultant disentangled representation. Moreover, InterFaceGAN\cite{shen2020interfacegan,saha2021loho,zhu2020domain} utilize StyleGAN's inversion methods to manipulate various facial attributes such as smile, hair, and glasses. Furthermore, some other methods \cite{lewis2021tryongan,zhu2020domain} finetune StyleGAN in non-face dataset and use inversion methods due to their purposes.

\section{The Proposed Framework}
\label{sec:Proposed_Framework}
\begin{figure*}
    \centering
    \includegraphics[width=\linewidth]{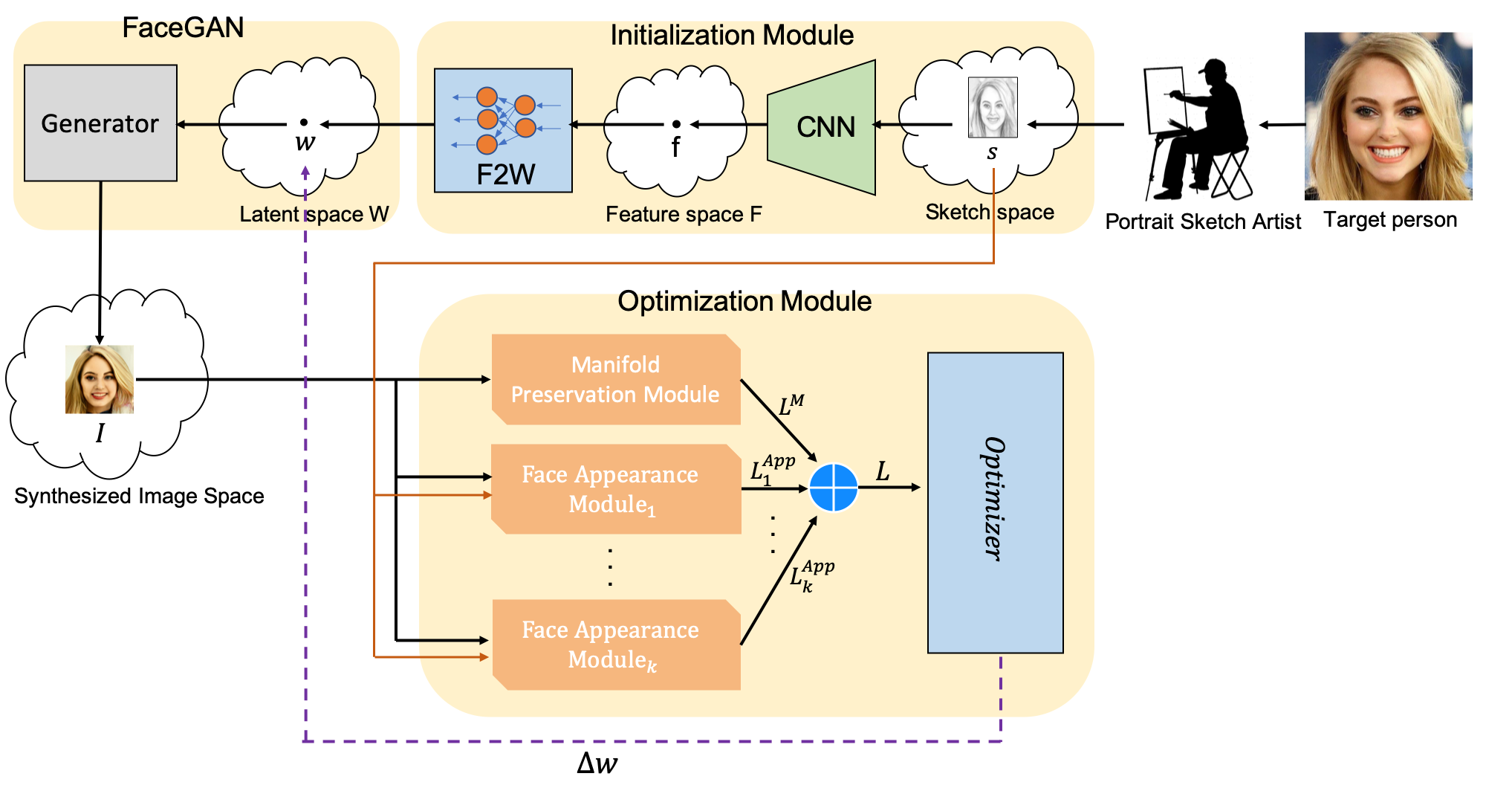}
    \caption{Block diagram of the proposed model.}
    \label{fig:block1}
\end{figure*}

Our proposed model to generate a realistic photo for a target person’s face is given in Fig. \ref{fig:block1}. As stated before, it is assumed that a sketch portrait of a person is provided and the aim is to convert this sketch to a photo-realistic image of the person. The overall process is organized into two main steps, namely, \emph{initialization} and \emph{fine-tuning} steps. The model, utilizes a face-generating sub-module mapping the latent space vectors, $w$, to real high-quality face images. The goal of the first step is to find an initial representation for the input sketch in the latent space of the face-generator module. This initial point is fed into the face-generator to create the initial image of the target person. The second step of the model is an optimization step that iteratively adjusts the initial image based on the high-level features extracted from the input sketch and the prior knowledge about realistic face images. The final output of this step is an artifact-free real face image similar to the input sketch. The details of all components of the proposed model are described in the following subsections.

\subsection{Face Generating Module}
One of the components of our proposed model is a generative model (\emph{FaceGAN}) with its own latent space which generates face images in high diversity and high quality. The face reconstruction task is then defined as finding the best match of the target face by searching the latent space of this face generator. The quality of the synthesized faces and their closeness to the target person is limited by the quality and diversity of the faces constructed by FaceGAN.

Any face generator with latent space $Z$ or intermediate latent space $W$ can play the role of the FaceGAN part of our model. There are lots of pre-trained models for this purpose. Progressive Growing GAN \cite{karras2017progressive} and Style-based Generative Adversarial Network (StyleGAN) \cite{karras2019style} are two such models.
StyleGAN is an extension to the GAN architecture that proposes large changes to the generator model, including the use of a mapping network to map points in the latent space to an intermediate latent space and the use of the intermediate latent space to control style at each point in the generator model. The resulting model not only does generate impressively photo-realistic high-quality photos of faces, but it also offers control over the style of the generated images through varying the style vectors. 

We have used StyleGAN in our proposed framework which consists of two main subnetworks to map $1 \times 512$ latent noise vectors to $18 \times 512$ intermediate latent codes ($W$) and the intermediate latent codes to realistic face images. Here, we directly work on the intermediate latent space as it is closer to the high-level features of face.  Our experiments show that for any given real face photo, the StyleGAN is capable of generating a similar image with very high fidelity. However, the large number of dimensions of the intermediate latent space and the locality of the changes make the search process complicated.

\subsection{Initialization Module}
\label{subsec:init_module}
In the initialization step of the proposed model, the input portrait sketch is mapped to the most similar vector in the intermediate latent space of FaceGAN. It would be ideal if the initialization step could implement the inverse function of FaceGAN, generating an intermediate latent vector corresponding to the photo-realistic image of the given sketch. As learning this reverse mapping as a single step is a very hard and complex task, we chose to implement the initialization module in two steps. In this setting, the input sketch is first translated into a high-level feature vector ($f$) and then this feature vector is mapped into the intermediate latent space of FaceGAN. The initialization method is trained to find the best estimation of the latent vector of the input sketch.

Regarding the feature extraction step of the initialization phase, our studies show that face recognition models such as FaceNet \cite{schroff2015facenet}, VGG16 \cite{simonyan2014very_VGG16}, VGGFace \cite{parkhi2015deep_VGGFACE}, and ResNet \cite{cao2018vggface2} are good choices for this task. While these models are trained on realistic photos, most of the features extracted by these models are color-independent and they can be successfully used to provide a high-level representation of the input colorless sketches. In this paper, we use VGGFace as it provided better performance in the experiments. See Section \ref{Exp:init_comp} for a comparison of the results of all feature extraction models.

A neural network (called F2W) is then trained to map the extracted feature vector to a vector in the latent space. To train F2W, a large number of images were randomly generated by FaceGAN. For each image, the $w$ vector was also recorded. The feature vectors of all images were extracted by VGGFace resulting in a data set of $(w,f)$  pairs. F2W is then trained with this data. We tested a number of different architectures for F2W. Finally, a simple fully connected network with two input and output layers was found to provide the best results.

\subsection{Optimization Module}
The initialization step generates a real face photo resembling the main face characteristics of the input sketch. After that, an iterative optimization procedure can be employed to improve the quality of the synthesized image by removing artifacts, capturing more structural features of the input sketch, and enforcing extra features like skin color. Here, we emphasize making the produced photo more similar to the corresponding sketch in terms of facial features while keeping it free from artifacts. However, the proposed model is completely flexible and other sub-goals can be added straightforwardly.

The optimization task of the proposed model is formulated as minimizing the overall loss function $L(w)$ where $w$ is the intermediate latent vector of the FaceGAN module and $L(w)$ is sum of a set of loss functions each corresponding to a sub-goal of the refinement step.

As stated before, in this paper, we consider diverse facial features capture and artifact removal sub-goals. As shown in the optimization module in Fig. \ref{fig:block1}, we have one manifold preservation module to eliminate the image artifacts and some face appearance modules to minimize the distance between various facial feature vectors of the sketch and the generated colorful image. So, the goal function of the model can be written as
\begin{equation}
\begin{aligned}
\label{equ:equ1}
    L(w)=L_1^{App}+...+L_k^{App}+L^M
\end{aligned}
\end{equation}
where $L_i^{App}$ denotes the loss related to the $i$th face appearance module and $L^M$ gives the loss of the manifold preservation module.

At each optimization step, the latent vector $w$ is updated using a gradient descent algorithm so that the generated image continuously gets closer to the input sketch.

Different parts of the optimization module of the proposed model are explained in the following subsections.

\subsubsection{Face Appearance Modules}

The most desired property of the synthesized face photo is to have facial features similar to those of the target person. To achieve this goal, we first extract the high-level feature vectors of the provided sketch and the generated image and then define the loss function as the discrepancy between these two vectors. Pre-trained face recognizers are good choices for facial feature extraction as their performance greatly depends on the feature sets they use. While most of these features are common in different models, the features extracted and learned by each face recognition model depend on the training dataset, training algorithms and the model's architecture. So, as our experimental results show, using a diverse set of feature vectors obtained from different face recognition models improves the quality of the generated faces.

\begin{figure}
    \centering
    \includegraphics[width=\linewidth]{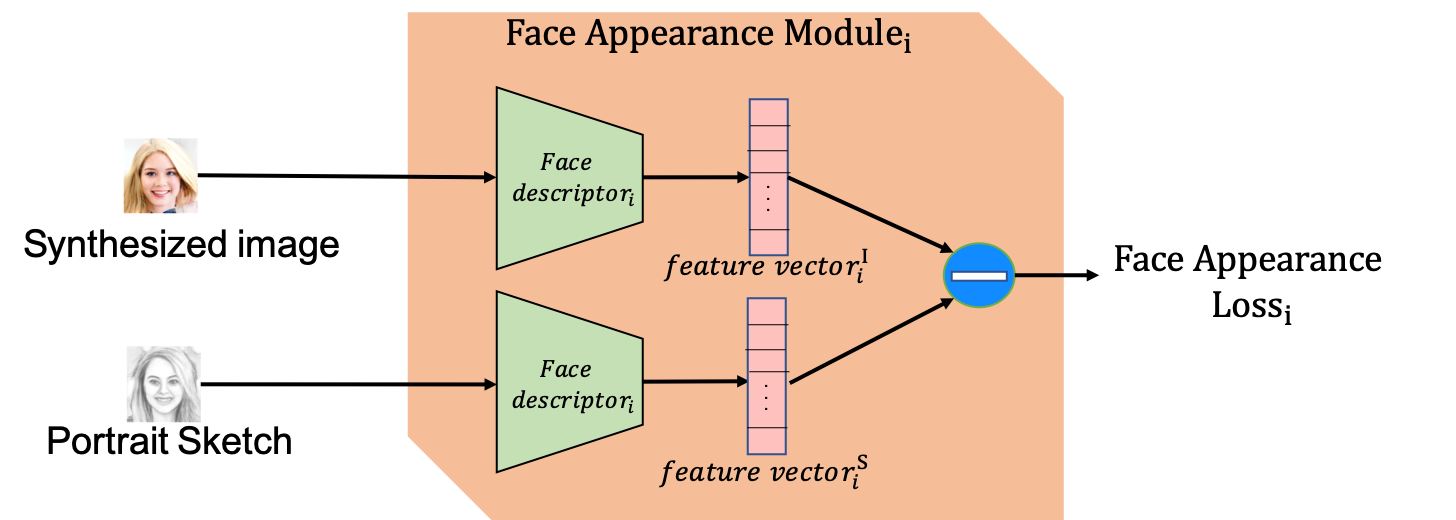}
    \caption{The structure of a face appearance module.}
    \label{fig:block2}
\end{figure}

As shown in Fig. \ref{fig:block2}, each feature extractor is applied to both the synthesized image and the sketch and the appearance loss is calculated as the Euclidean distance of the corresponding feature vectors as follows
\begin{equation}
\label{equ:equ2}
\begin{aligned}
    L_i^{App}=||FR_i(syn)-FR_i(sketch)||_2\\
\end{aligned}
\end{equation}
where $FR_i(I)$ is the feature vector of image $I$ given by the $i$th feature extractor.

In this paper, we have studied the feature vectors of some state-of-the-art face recognition models including VGGFACE \cite{parkhi2015deep_VGGFACE} and VGGFACE2 \cite{cao2018vggface2} as well as classification networks like VGG16 \cite{simonyan2014very_VGG16}. The results are reported and discussed in Section \ref{Exp:opt_comp}.

\subsubsection{Manifold Preservation Module}

The goal of the FaceGAN is to map an intermediate latent vector $w$ to a high-resolution realistic face photo. However, not all points of this latent space are converted to clear face images; many input points result in faces with different kinds of artifacts or even completely non-face images. The non-face photos are penalized by the appearance modules as their feature vectors are far from those of the input sketches but it is still possible to have face photos with frequent small artifacts. So, we have added another module to our proposed model to keep $w$s of the low-dimensional subspace of the latent space whose points produce artifact-free face photos and hence we call it the \emph{manifold-preservation} module. As the experimental results show, the vectors on this manifold generate faces with proper structural and appearance characteristics while the artifacts of the image increase as we get away from this manifold. 

\begin{figure}
    \centering
    \includegraphics[width=\linewidth]{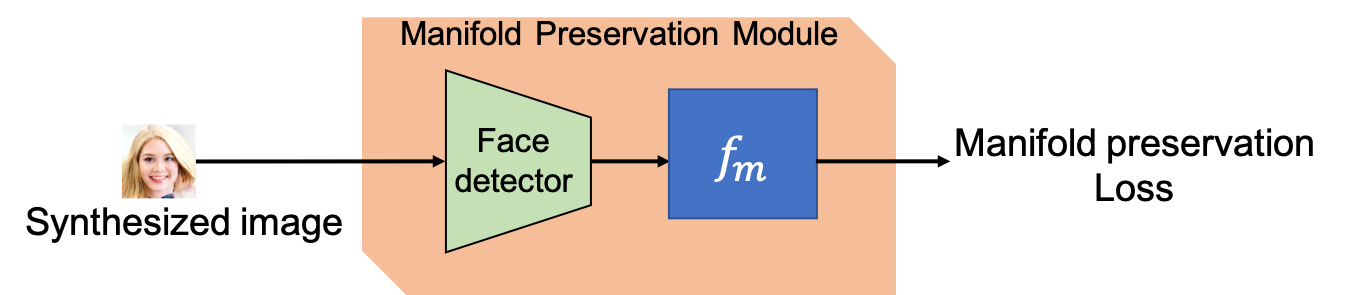}
    \caption{Manifold preservation module}
    \label{fig:block3}
\end{figure}

As stated before, we need to quantify the amount of artifacts present in a produced image. Here, we train a convolutional neural network, called HOGFD, which generates \emph{faceness} scores for each synthesized image. This score evaluates the goodness of the image based on its closeness to a real face photo. The output of HOGFD is converted to the manifold-preservation loss as follows:
\begin{equation}
\label{equ:equ3}
\begin{aligned}
    L^{M}(syn)=max\_score-HOGFD(syn)\\
\end{aligned}
\end{equation}
where $syn$ denotes the synthesized image and $max\_score$ is the highest scores of $HOGFD$ network for the set of training face images. The structure of the proposed manifold preservation module is depicted in Fig. \ref{fig:block3}. The $f_m$ in this figure shows the mathematical subtraction stated in Eqn. \ref{equ:equ3}.

HOGFD network is trained using the HOG-based face detector proposed by \cite{dalal2005histograms} and \cite{dlib09}. To this end, a hundred thousand of generated images by FaceGAN are selected. The output of the HOG-based face detector is calculated for all these images as the faceness score and then HOGFD network is trained to estimate this score for the input images. 
The HOGFD network has four 3 x 3 convolutional layers with 16, 32, 64, and 128 filters and stride 1, each followed by a max pooling layer. Three fully connected layers with 16, 4, and one neuron exist on the top of the network to do the regression task. Batch normalization is used after each convolution layer and first fully connected layer to make the network faster and more stable. A dropout layer with probability 0.5 is used to avoid overfitting and Relu activation used as non-linear activation function for all layers.

\section{Experiments}
\label{sec:Experiments}

In this section, we conduct extensive experiments to verify the effectiveness of the proposed method. We compare our method qualitatively and quantitatively with state-of-the-art synthesis methods. We use the AR dataset \cite{martinez1998ar} part of the popular CUHK Face Sketch Database (CUFS), which contains 123 images including 70 males and 53 females in our tests. All photos of this dataset are captured under well-controlled conditions along with the artist-drawn sketch of each photo.  \cite{wang2008face}. 

\subsection{Component Analysis}
The experiments of this section investigate the role of each component of our proposed framework. At first, the power of the initialization component (F2W network) in mapping the initial image to a vector in the latent space with similar face characteristics is studied. After that, the results of the optimization step with different loss functions are presented and discussed. Finally, the scores that HOGFD network assigns to a number of face photos with different qualities are reported to show the important role of this component in guiding the search toward high-quality faces.

\subsubsection{Initialization Component}
\label{Exp:init_comp}
\begin{figure}
    \centering
    \includegraphics[width=\linewidth]{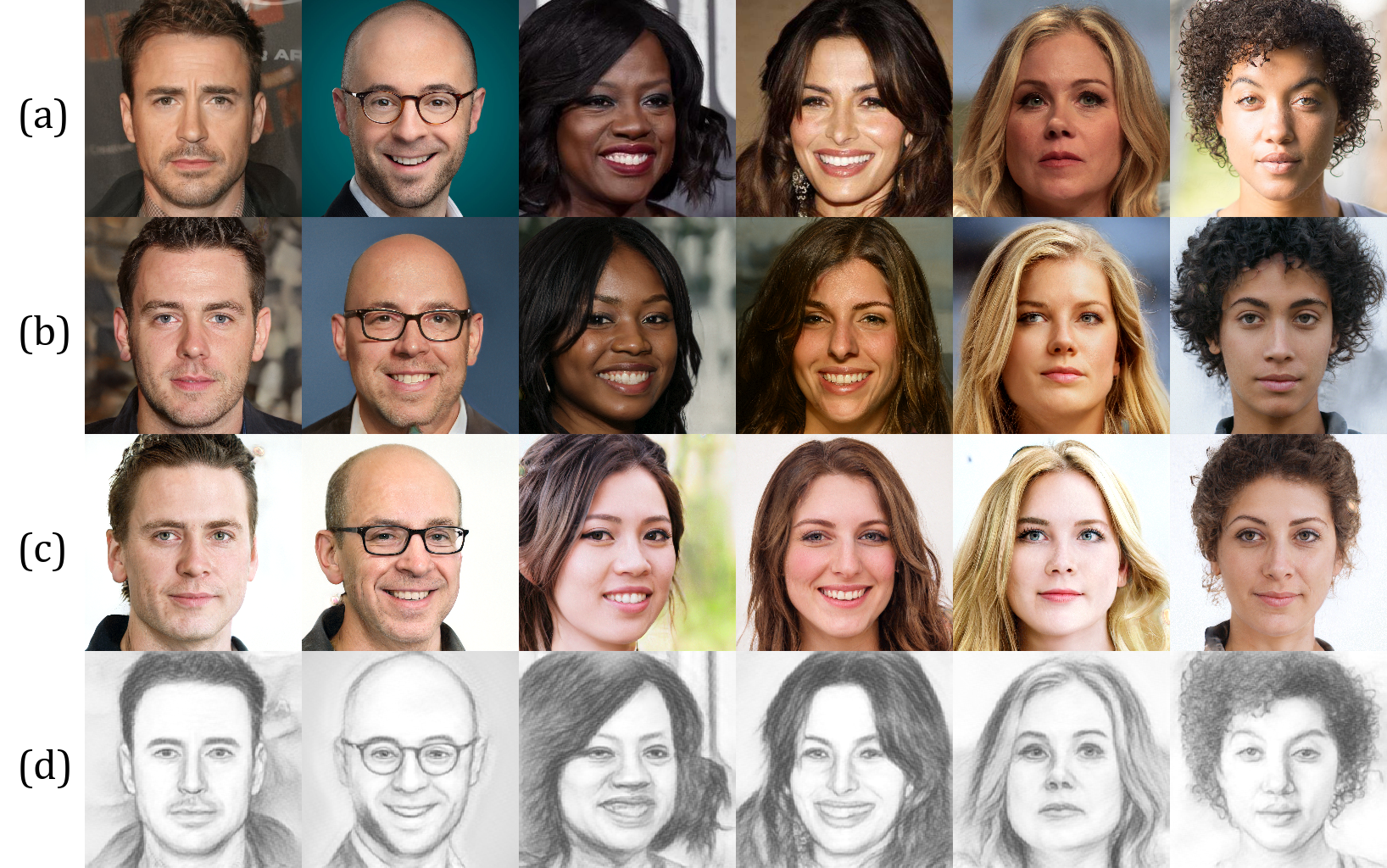}
    \caption{Initialization module and F2W network: (a) initial colored images, (d) corresponding sketches (b) result of F2W network applied to the colored images, (c) result of F2W network applied to the sketches. }
    \label{fig:F2W_Res}
\end{figure}

In Fig. \ref{fig:F2W_Res}, the F2W network is applied to six colored images and their corresponding sketches obtained by \cite{wang2018high} model. The sketches shown in row (d) are corresponding to the colored images of row (a). Rows (b) and (c) respectively show the results of mapping the colored images and sketches. The images are first mapped into the latent space of the FaceGAN module and then the corresponding photos are generated by FaceGAN. As can be seen in this figure, the mapping preserves the high-level features (like sex, mouth state, hair type) of the faces in both cases. However, there are still noticeable differences that are addressed in the optimization step. 

\begin{figure}[hbt!]
    \centering
    \includegraphics[width=\linewidth]{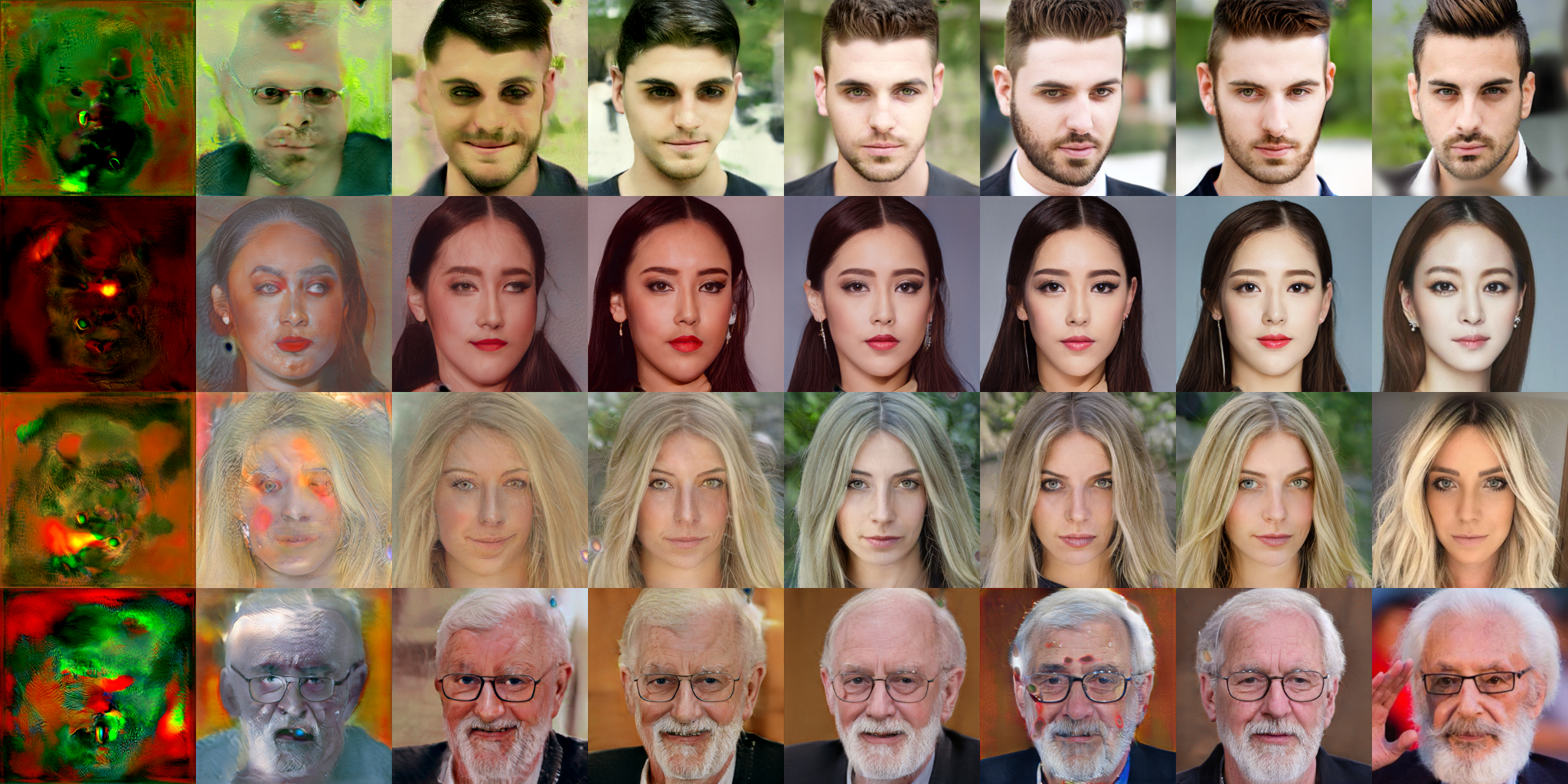}\\
    \begin{tabular}{p{0.035\textwidth}p{0.035\textwidth}p{0.035\textwidth}p{0.037\textwidth}p{0.037\textwidth}p{0.037\textwidth}p{0.037\textwidth}p{0.037\textwidth}}
        20 & 40 & 80 & 160 & 320 & 620 & 1000 & input \\
    \end{tabular}
    \caption{F2W network's improvement during training; numbers show the training steps. }
    \label{fig:F2W_stepbystep}
\end{figure}

The above performance is achieved by training the F2W network on a large dataset of $(f,w)$ pairs where $w$ is the latent vector and $f$ is the feature vector of the face image generated by FaceGAN from $w$. Fig. \ref{fig:F2W_stepbystep} shows the progress of the F2W network during its training. In this figure, the mapping results of the F2W network for a set of test images are shown after 20, 40, 80, 160, 320, 620, and 1000 steps of training. It is clear how the network’s output moves from noisy meaningless images at the early stages to images very similar to the input images at the final stages. 

\subsubsection{Optimization Process}
\label{Exp:opt_comp}
\begin{figure}[hbt!]
    \centering
    \includegraphics[width=\linewidth]{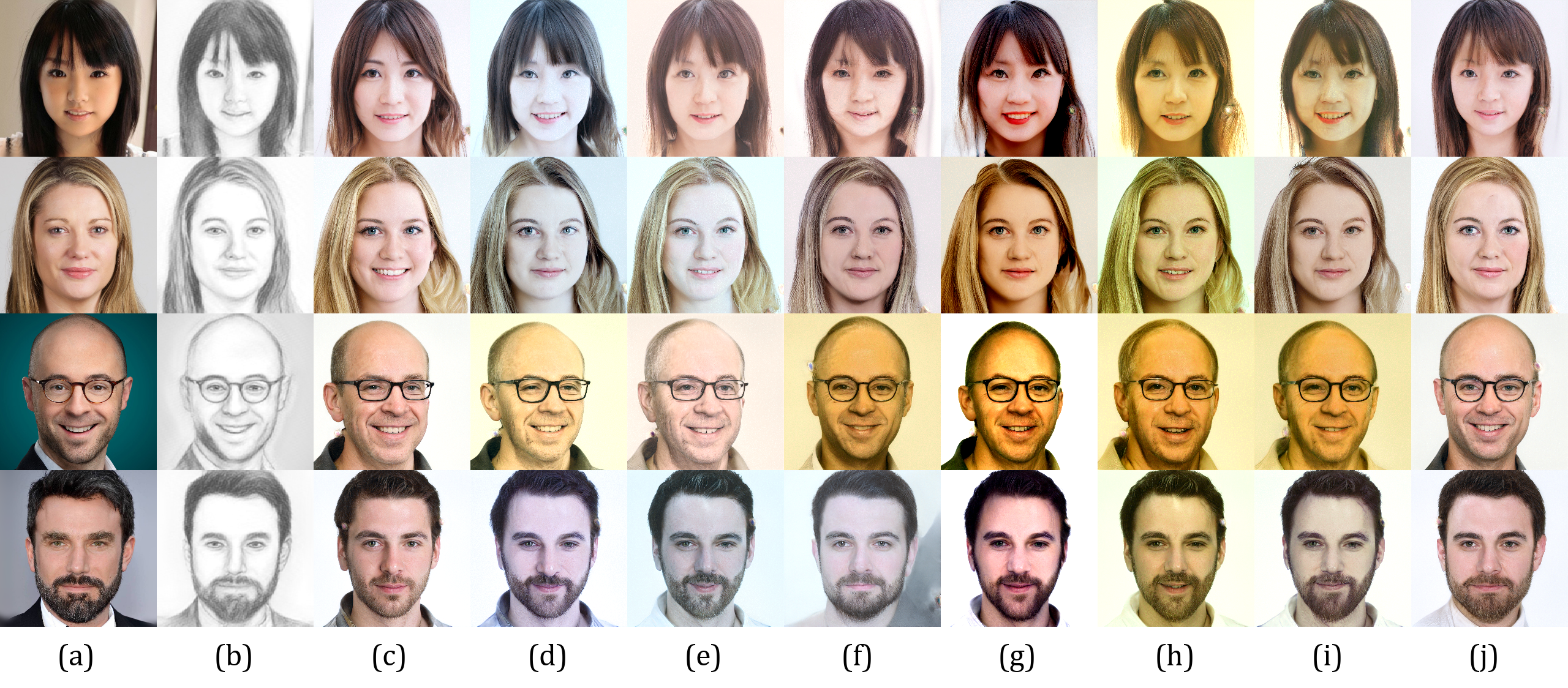}
    \caption{Ablation study of the proposed loss function. The input image, its sketch, and the output of F2W for sketch are shown in columns (a), (b) and (c), respectively. Columns (d) to (j) show the results of different loss functions:, (d) $L_1^{App}$, (e) $L_2^{App}$, (f) $L_1^{App}$ + $L_2^{App}$ + $L_3^{App}$, (g) $L_1^{App}$ + $L_M$, (h) $L_2^{App}$ + $L^M$, (i) $L_1^{App}$ + $L_2^{App}$ + $L^M$, (j) $L_1^{App}$ + $L_2^{App}$ + $L_3^{App}$ + $L^M$. }
    \label{fig:lossCompare}
\end{figure}

To show the contribution of each part of our proposed loss function to the final model, we have performed the optimization step with different combinations of loss functions. Here, we test four loss functions: appearance loss of VGGFace ($L_1^{app}$), appearance loss of VGGFace2 ($L_2^{app}$), manifold preservation loss ($L_M$) and feature loss of VGG16 ($L_3^{app}$). Fig. \ref{fig:lossCompare} shows the results of the ablation study of the proposed loss function. As the results show, the best performance is achieved by the combination of all loss functions. But the identities of the generated face photos by the proposed method still have a little difference with the real photos which is mainly due to the low quality of the input sketches.

\subsubsection{Manifold Preservation Scores}
\begin{figure}
    \centering
    \begin{tabular}{ccccc}
        \includegraphics[width=24mm]{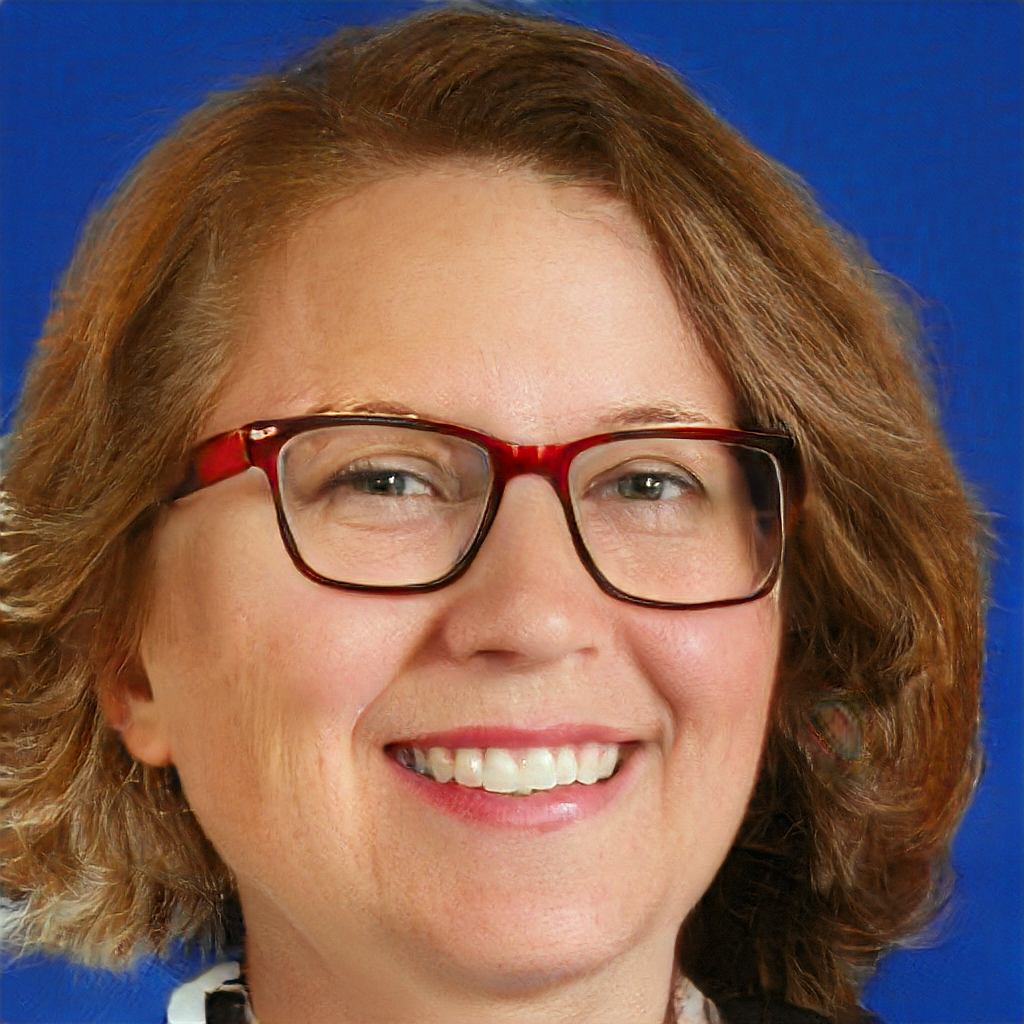} & \includegraphics[width=24mm]{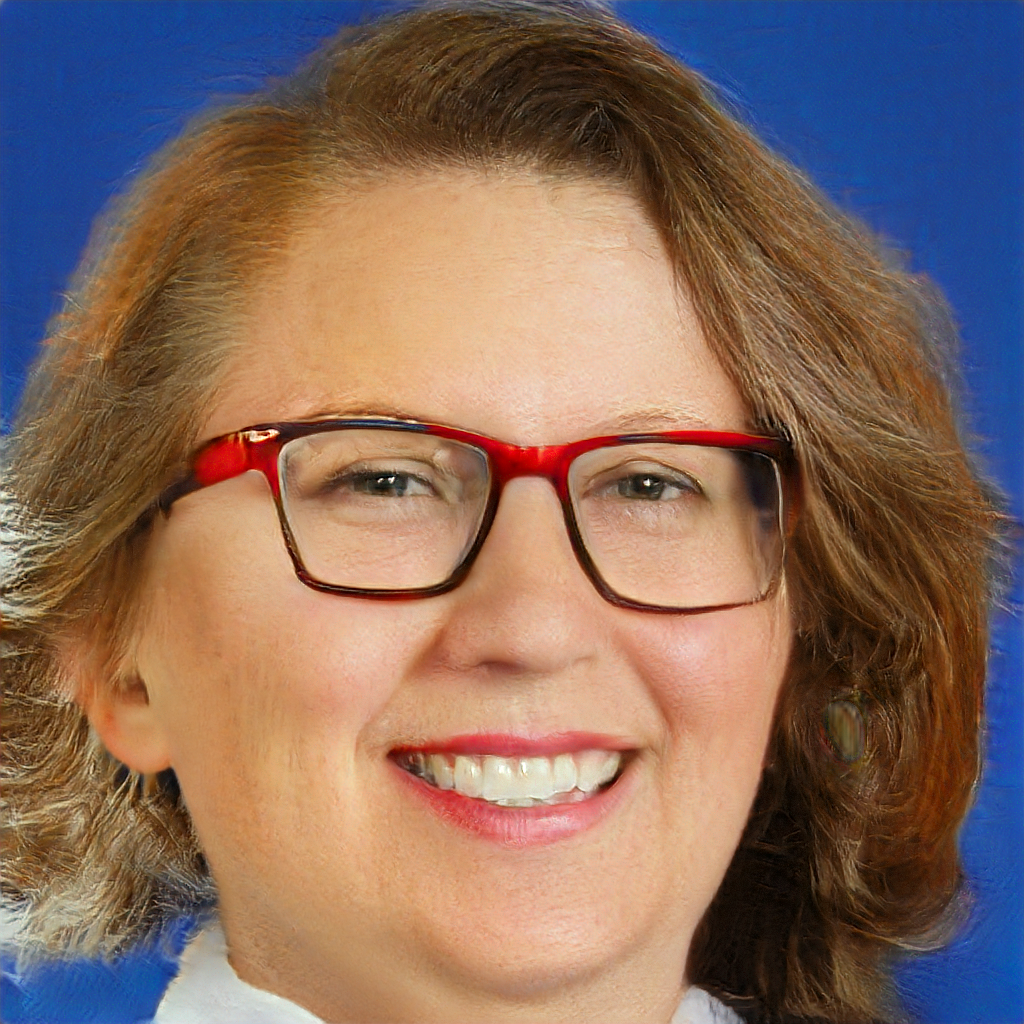} & \includegraphics[width=24mm]{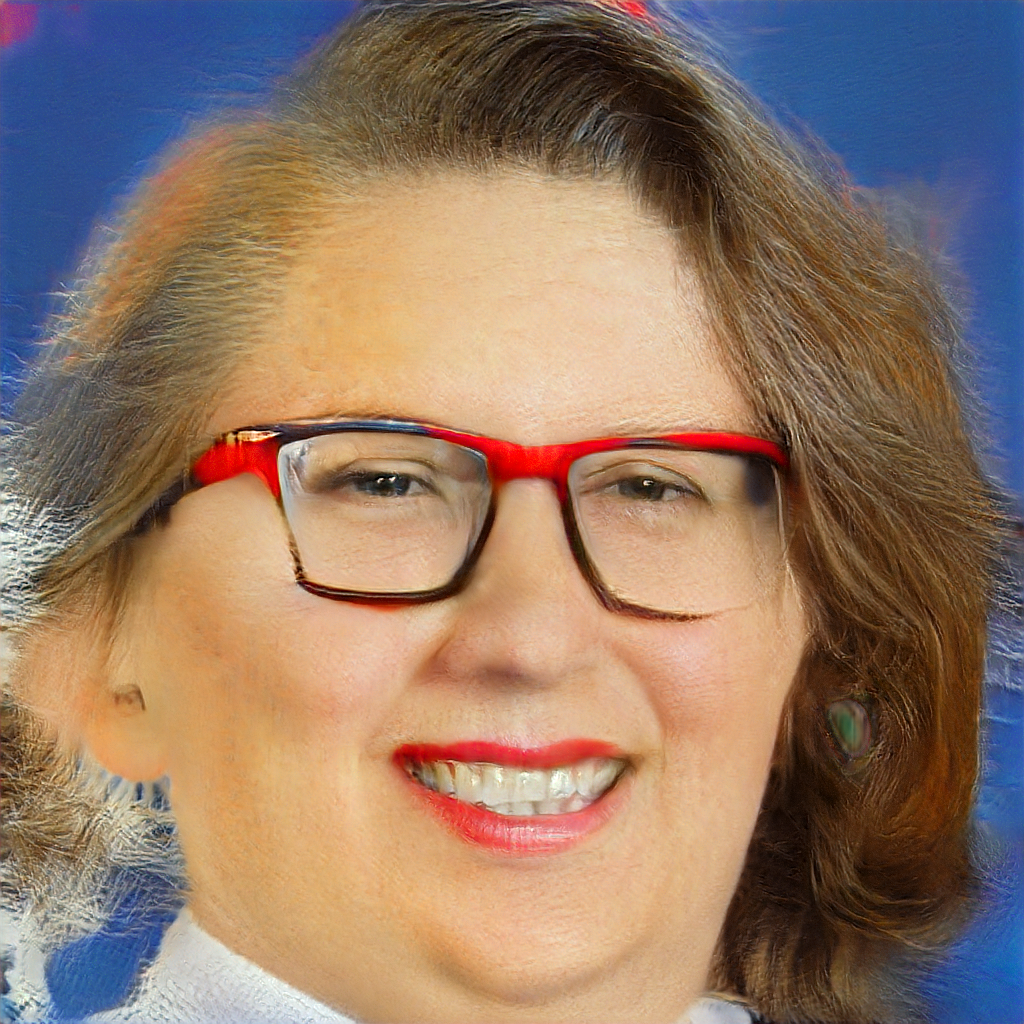} &
        \includegraphics[width=24mm]{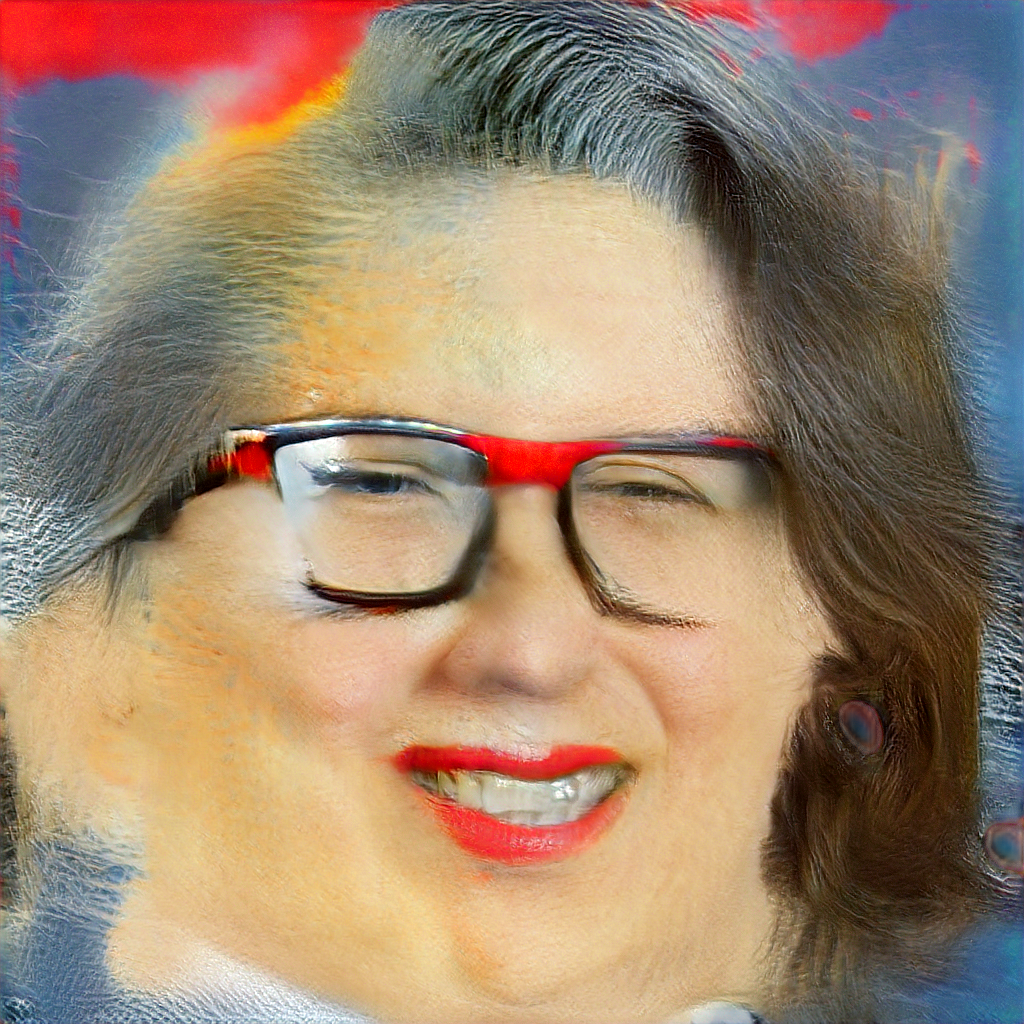} &   
        \includegraphics[width=24 mm]{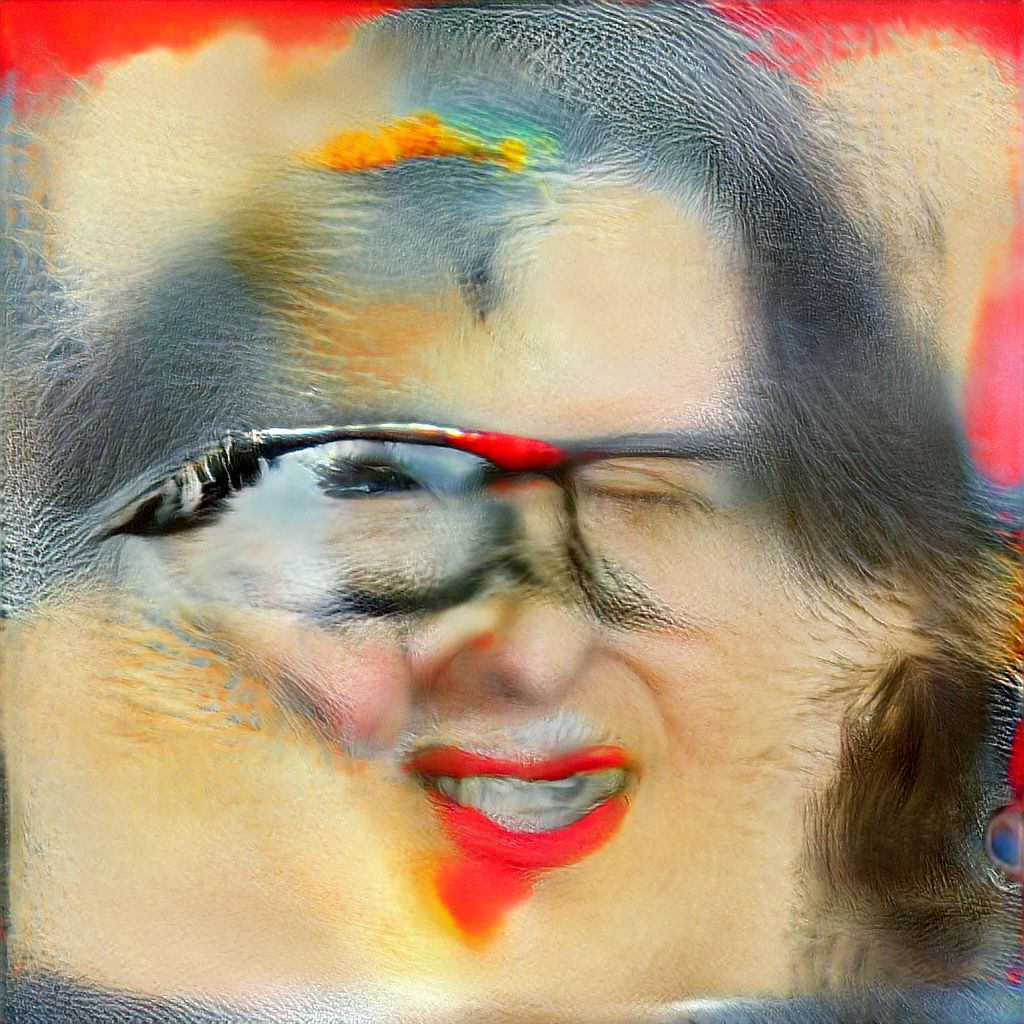} \\
        (a)1.84 & (b)1.79 & (c)1.77 & (e)1.40 & (f)0.66 \\[6pt]
    \end{tabular}
    \caption{Manifold preservation scores.}
    \label{fig:artifact_loss}
\end{figure}

As mentioned before, not all intermediate latent vectors ($w$s) are converted to realistic face photos by FaceGAN. Desired $w$s form a small subspace of the whole space. We employed the HOGFD network to measure the distance of each $w$ vector from this manifold. To this end, HOGFD assigns a score to each face image based on the amount of artifact presented in that image. Fig. \ref{fig:artifact_loss} shows samples of these scores for some images produced by StyleGAN for different $w$ vectors. The HOGFD's score for an image is inversely proportional to the image quality degradation. 

\subsection{Comparison with Other Methods}
\begin{figure}[hbt!]
    \centering
    \includegraphics[width=\linewidth]{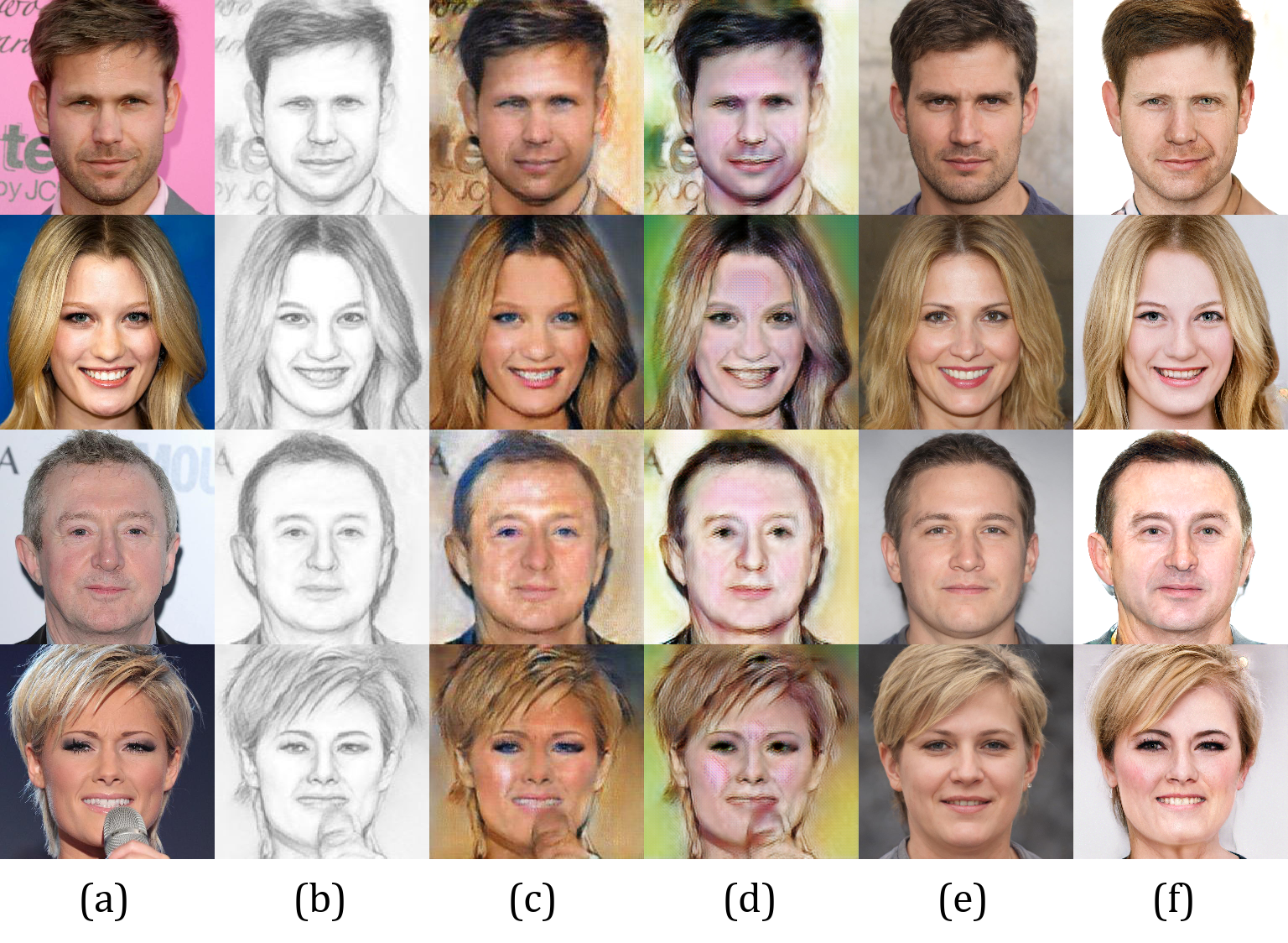}
    \caption{The results of photo to sketch models on Celeba-HQ dataset: (a) colored images, (b) sketches, (c) DualGAN’s results, (d) CycleGAN’s results, (e) pixel2style2pixel’s results, and (f) our proposed model’s results.}
    \label{fig:compareWithOthers}
\end{figure}

In this section, the results of our proposed model are compared with those of four other methods. In the first experiment, one thousand images from the Celeba-HQ dataset \cite{karras2017progressive} are selected and divided into 900 training and 100 test images. These images are then used to train DualGAN \cite{yi2017dualgan}, CycleGAN \cite{zhu2017unpaired}, and fine-tune pixel2style2pixel \cite{richardson2021encoding} sketch to photo models. Fig. \ref{fig:compareWithOthers} shows the results of these models along with the photos generated by our proposed model for some test images. The DualGAN and CycleGAN models stick to the input sketch and its background and just add almost the same skin color to all sketches while our proposed model reconstructs the face from the scratch, keeps all details of the face parts, and add no artifact to it. As one can see, the background and even the overlapping objects like the microphone in the last row are removed by the proposed model. However, the pixel2style2pixel model which is based on the StyleGAN face generator generates high-quality images comparing with two other models, while the generated images are still far from the target in identity aspect. Moreover, one considering point is that there is no need to train the proposed model on the new dataset, while other models are trained on this used dataset.

\begin{figure}[hbt!]
    \centering
    \includegraphics[width=\linewidth]{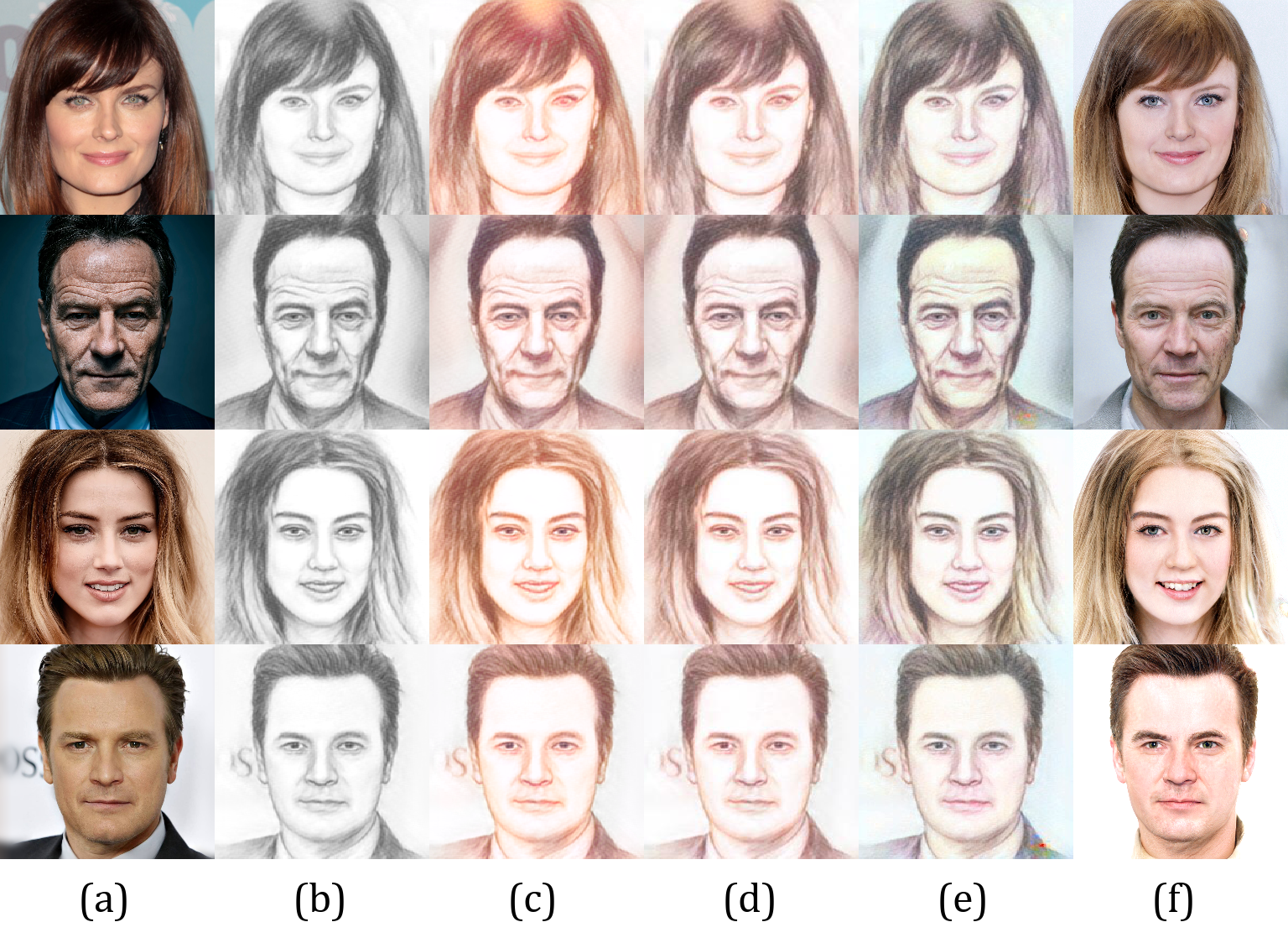}
    \caption{The results of the proposed model and some online image colorizing applications: (a) colored images, (b) sketches, (c) hotpot’s results, (d) imagecolorizer's results,(e) Photomyne’s results, and (f) the proposed method’s results.}
    \label{fig:compareWithApps}
\end{figure}

In another experiment, we compare our proposed model with three online black and white image colorizing applications: hotpot \cite{hotpot}, imagecolorizer \cite{imagecolorizer}, and Photomyne \cite{Photomyne}. The results are shown in Fig. \ref{fig:compareWithApps}. It is clear from the results that the proposed model is far superior to the compared web-apps. All applications slightly change the color of the lines of the sketch to a reddish-brown color while the proposed model gives fully-colored photos and even properly guesses some hair color.

\subsection{Quantitative Evaluations}
We used two sets of criteria to evaluate the proposed method quantitatively, 1) full-reference image quality assessment and 2) rank-1 face recognition accuracy measures. In full-reference image quality assessment, the main real image is considered as the reference image and the output image of the proposed model is compared with it. Here, we use three image quality assessment measures: 1) Structural Similarity Index (SSIM), 2) Feature Similarity Index (FSIM), and 3) Visual Information Fidelity Index (VIF). As mentioned earlier, part of the AR dataset is used in this test; 40 pictures are randomly selected for training CycleGAN and DualGAN methods, and fine-tuning the pixel2style2pixel model and the other 83 pictures are used for the test. The results are given in table \ref{tab:SSIM_FSIM_VIF}. SSIM measure quantifies the perceived quality of an image and as the results show, it is much higher in the images produced by the proposed model compared to CycleGAN and DualGAN. This is due to the optimization process of the proposed model that prevents output image from being blurred or noisy. Also, the VGG16 loss function leads to a similar structure of the sketch and output image. For the case of VIF metric that correlates with a human judgment of image quality the proposed model performs better than CycleGAN and DualGAN. Regarding the FSIM which is more related to the local features of the two images, the CycleGAN and DualGAN perform better as they directly use the sketch in producing the final results while the proposed model works on the $w$ vectors in the intermediate latent space. On the other hand, the Pixel2style2pixel model by producing natural photos based on StyleGAN could gain higher scores in all of these metrics. But despite the high results of this model in this section, the identities of the produced photos are very far from the targets and this will be studied in the following. 

\begin{table}[!htbp]
\small
\centering
\caption{Full-reference image quality assessment results on part of the CUFS dataset. Here the proposed method is compared with three of existing advanced methods, including DualGAN \cite{yi2017dualgan}, CycleGAN \cite{zhu2017unpaired}, and pixel2style2pixel\cite{richardson2021encoding}.}
\label{tab:SSIM_FSIM_VIF}
\addtolength{\tabcolsep}{-4.1pt}
\begin{tabular}{p{2.3cm}p{1.9cm}p{1.9cm}p{1.9cm}}
\toprule
\multirow{2}[3]{*}{Method} & \multicolumn{3}{c}{Criterion}  \\
\cmidrule(lr){2-4} 
 & SSIM & FSIM & VIF \\
\midrule
DualGAN & 0.57 & 0.755 & 0.109 \\
CycleGAN & 0.57 & 0.754 & 0.108 \\
Pixel2style2pixel & \textbf{0.79} & \textbf{0.781} & \textbf{0.147} \\
Proposed model & 0.655 & 0.748 & 0.115  \\
\bottomrule
\end{tabular}
\end{table}

In the other test of this section, the generated photos are used for face recognition (FR). The recognition results are reported in Table \ref{tab:Rank-1}. Three different networks (VGGFACE, VGGFACE2, and FaceNet) are used for face recognition. As the results show, the recognition accuracy for the images synthesized by the proposed model is higher than the other methods. For VGGFACE and VGGFACE2 whose features are included in the optimization process, the superiority of the proposed model is quite clear and for the FaceNet which is not considered in the optimization, the performances of models are close to each other indicating that the proposed model can generate realistic images without degrading identity. Moreover, the low accuracy of the pixel2style2pixel is due to the difference between the identity of the generated photos with their corresponding targets.

\begin{table}[!htbp]
\small
\centering
\caption[align=c]{Rank-1 face recognition accuracies on part of the CUFS dataset. Here the proposed method is compared with three of existing advanced methods, including DualGAN \cite{yi2017dualgan}, CycleGAN \cite{zhu2017unpaired}, and pixel2style2pixel\cite{richardson2021encoding}. Also three different networks are used for face recognition, including VGGFACE, VGGFACE2, and FaceNet.}
\label{tab:Rank-1}
\addtolength{\tabcolsep}{-4.1pt}
\begin{tabular}{p{2.3cm}p{1.9cm}p{1.9cm}p{1.9cm}}
\toprule
\multirow{2}[2]{*}{Method} & \multicolumn{3}{c}{FR Method}  \\
\cmidrule(lr){2-4} 
 & FaceNet & VGGFace & VGGFace2 \\
\midrule
DualGAN & \textbf{0.506} & 0.855 & 0.650 \\
CycleGAN & 0.481 & 0.795 & 0.698 \\
Pixel2style2pixel & 0.085 & 0.158 & 0.098 \\
Proposed model & \textbf{0.506} & \textbf{0.975} & \textbf{0.831} \\
\bottomrule
\end{tabular}
\end{table}

\subsection{Qualitative Evaluation}
For subjective evaluation of the results by real users, we created a form containing 50 sketches and the corresponding colorful images created by different sketch-to-photo methods including CycleGAN, DualGAN, pixel2style2pixel(pSp), and our proposed method. Users were asked to select the best real photo for each sketch based on two main criteria: 1) does the generated image looks realistic? and 2) do the sketch and generated image represent the same identity? To avoid biasing users towards the results of a special method, the results are shuffled for each sketch. The images are selected from the AR dataset and CelebA dataset. Sixty people have participated in this evaluation. The percentage of times each method has been selected as the best method is reported in Table \ref{tab:user_study}. As the results show the proposed method outperforms the other methods from the users' perspective.

\begin{table}[!htbp]
\small
\centering
\caption[align=c]{Result of the subjective evaluation. The results of the perceptive user study conducted on 60 people over 50 samples including sketches and generated photos by four methods. }
\label{tab:user_study}
\addtolength{\tabcolsep}{-4.1pt}
\begin{tabular}{p{2.4cm}p{1.6cm}p{1.6cm}p{1.2cm}p{1.2cm}}
\toprule
Method & DualGAN & CycleGAN & pSp & Our's \\
\midrule
success percent & 8.03\% & 4.35\% & 11.72\% & \textbf{75.89\%} \\
\bottomrule
\end{tabular}
\end{table}

\subsection{Complexity Analysis}
There are two main categories of solutions for the sketch to photo translation task. The first category relies on encoder-decoder architectures. These models need a time-consuming training process to train the model parameters, while they are fast in the test as the final output is generated by only one pass through the encoder and decoder networks. The second category is based on optimization in which a pre-trained target image generator is utilized along with some objective functions. These models don't need a training phase and training dataset. However, the execution time of these models is generally higher than in the first category. Our proposed model belongs to the second category of models and is a general framework without any training phase for a new dataset. The running time of our model on a Ubuntu 20.04 system with 16 GB RAM and one 1080 Ti GPU is around 3-5 minutes per image. The models of the first category need a training phase that continues for several days and test runtime of few seconds.

\section{Conclusions And Future Works}
\label{sec:conclusion}
In this paper, we proposed a novel sketch to photo synthesis framework based on GANs. Unlike most of the existing methods, the proposed method requires no paired data and no training for new datasets. The overall process is divided into an initialization step which maps the input sketch to the intermediate space of the face synthesizer and the optimization step which fine-tunes the generated face photo. The proposed framework is quite flexible and can be easily adapted to any face synthesizing model and any fidelity criterion. The proposed model was applied to several sketches from different datasets and the results were evaluated based on the objective and subjective measures. As the experiments show, the output of the model is a high-quality and realistic face photo whose attributes are the same as the input sketch. Furthermore, the high recognition rates over the synthesized images show the superiority of the proposed model in preserving the identity of the target person.

In the following, the proposed model can be extended by incorporating attributes not provided by the sketch, e.g. skin or hair color, either directly from an input vector or indirectly from a face photo.

\bibliographystyle{unsrt} 
\bibliography{references}

\end{document}